\documentclass{iLRNConfProc}
\usepackage{subcaption}
\usepackage{listings}

\title{Understanding Generalization, Robustness, and Interpretability in Low-Capacity Neural Networks}

\author{
\begin{minipage}{\textwidth}
\centering
Yash Kumar, Indian Institute of Technology Madras \\
\texttt{hi@yashkc.com}, \texttt{22f3000472@ds.study.iitm.ac.in}
\end{minipage}
}

\date{}

\begin{document}
\maketitle
\thispagestyle{empty}

\vspace{3\baselineskip}

\section*{Abstract}
Although modern deep learning often relies on massive over-parameterized models, the fundamental interplay between capacity, sparsity, and robustness in low-capacity networks remains a vital area of study\cite{ref11}. We introduce a controlled framework to investigate these properties by creating a suite of binary classification tasks from the MNIST dataset with increasing visual difficulty (e.g., \texttt{0 and 1} vs. \texttt{4 and 9}). Our experiments reveal three core findings. First, the minimum model capacity required for successful generalization scales directly with task complexity. Second, these trained networks are robust to \textbf{extreme magnitude pruning (up to 95\% sparsity)}, revealing the existence of sparse, high-performing subnetworks. Third, we show that over-parameterization provides a significant advantage in \textbf{robustness against input corruption}. Interpretability analysis via saliency maps further confirms that these identified sparse subnetworks preserve the core reasoning process of the original dense models. This work provides a clear, empirical demonstration of the foundational trade-offs governing simple neural networks.

\section{Introduction}
The remarkable success of modern artificial intelligence is largely driven by the scaling of neural networks to massive sizes. Massive models like Transformers and deep convolutional networks have pushed the boundaries of performance across many challenging tasks. This progress often comes at the cost of heavy computational resources, consume substantial energy, and often operate as black boxes. Their lack of transparency makes them less suitable for scenarios where resources are limited or where understanding the model's reasoning is critical, especially given that their predictions can be vulnerable to imperceptible perturbations.\cite{ref14}

However, smaller networks can serve as an effective and efficient alternative. They are essential for efficient deployment on edge devices and provide a tractable laboratory to develop a fundamental scientific understanding of how neural networks learn. While the trade-offs between a model's capacity, the sparsity of its learned connections, and its robustness to corruption are often studied in complex settings, a clear, controlled analysis of these foundational properties is necessary.

In this work, we introduce a controlled framework for systematically investigating these relationships. Our key methodological contribution is the creation of a suite of binary classification tasks from the MNIST dataset where task difficulty is precisely modulated by selecting digit pairs of increasing visual similarity (e.g., the simple \texttt{0 and 1} task versus the challenging \texttt{4 and 9} task). This allows us to isolate the effect of task complexity on model behavior.

Our results reveal a clear and direct link between task difficulty and the minimum model capacity required for successful generalization. We demonstrate that these networks are extremely robust to magnitude pruning and show that over-parameterization, while not always necessary for clean data, provides a significant advantage in robustness against input noise and occlusion. We first describe our methodology, then present the results of our capacity, pruning, and robustness experiments, and conclude with a discussion of the findings.

\section{Related Work}
Our research connects two essential areas in deep learning: the pruning of networks and the examination of model capacity.

The search for more efficient neural networks has led to extensive research in network pruning, with modern compression techniques demonstrating the ability to remove a majority of parameters with minimal performance loss\cite{ref15}. A seminal contribution in this area is the \textbf{Lottery Ticket Hypothesis} by Frankle and Carbin (2019)\cite{ref4}, which proposes that dense, randomly initialized networks contain sparse subnetworks \textbf{winning tickets}. These subnetworks, if identified at initialization, can be trained in isolation to match or exceed the performance of the full, dense model. The LTH's methodology focuses on this "rewind-and-retrain" process to find promising subnetworks before training begins. In contrast, our work focuses on the properties of sparse subnetworks found via one-shot magnitude pruning after the dense model has been fully trained. We analyze the performance, internal representations, and reasoning processes of these emergent sparse structures without retraining, offering a complementary perspective on network sparsity.

In the same manner, recognizing the relationship between model capacity, generalization, and robustness is a key focus in deep learning. Much of the existing literature centers on the behavior and surprising efficiency of large, overparameterized models. In contrast, our contribution to this field lies in providing a carefully controlled and minimalist framework. By systematically varying the difficulty of tasks through visually similar digit pairs, we establish a clear and interpretable environment to analyze these foundational trade-offs, demonstrating the direct scaling of necessary capacity with the complexity of tasks.

\section{Methodology}

We base our experiments on the widely used \textbf{MNIST}\cite{ref1} dataset of handwritten digits. Rather than tackling the full 10-class classification problem, we designed a series of \textbf{binary classification tasks} by selecting digit pairs that vary in visual similarity. This method allowed us to control the complexity directly. The five digit pairs, ordered from easiest to most challenging, are: \texttt{0 and 1}, \texttt{1 and 7}, \texttt{5 and 6}, \texttt{3 and 8}, and \texttt{4 and 9}. For every task, we split the data into training and validation sets and repeated the entire process using three different random seeds to ensure the consistency of our results.

\subsection{Model Architecture and Training}

Our models are simple, fully connected neural networks. Each model consists of:

\begin{enumerate}
    \item An \textbf{input layer} with 784 units, representing the flattened 28x28 MNIST images.
    \item A single \textbf{hidden layer} with a variable number of neurons ranging from 2 to 64 units. This hidden layer serves as our main lever for adjusting model capacity. We use the \textbf{ReLU}\cite{ref3} activation function here.
    \item An \textbf{output layer} with one neuron and a \textbf{sigmoid} activation, yielding a probability score for the binary output.
\end{enumerate}

To tune the learning rate, we performed a grid search on the most difficult task (\texttt{4 and 9}) across all hidden layer sizes. The search covered learning rates from \(1 \times 10^{-4}\), \(3 \times 10^{-4}\), \(1 \times 10^{-3}\), \(3 \times 10^{-3}\), \(1 \times 10^{-2}\), \(3 \times 10^{-2}\). All models were trained for 100 epochs using \textbf{Stochastic Gradient Descent (SGD)} with \textbf{Binary Cross-Entropy} as the loss function.

\subsection{Experimental Protocol}

Our experiments were structured in four main phases:

\begin{enumerate}
    \item \textbf{Capacity Analysis:} We trained networks across the full range of hidden sizes on each of the five binary tasks to determine the minimum model capacity required for strong performance, and to observe how this threshold shifts with increasing task difficulty.

    \item \textbf{Sparsity Analysis:} After full training, we applied \textbf{one-shot magnitude pruning} at varying sparsity levels\cite{ref5}, up to 99\% to assess how well the models can retain performance when significantly compressed. This allowed us to test for the presence of highly sparse yet effective subnetworks.

    \item \textbf{Robustness Analysis:} For hidden sizes of \texttt{24} and \texttt{64}, we evaluated how models respond to input noise. Two types of corruption were introduced: (i) \textbf{additive Gaussian noise} with increasing variance, and (ii) \textbf{random occlusion}, where a $7\times7$ patch of pixels was masked out.\cite{ref13}

    \item \textbf{Interpretability Analysis:} To gain insight into what the models learn, we applied \textbf{t-SNE} to the hidden layer activations\cite{ref6} of the size = 24 network trained on the \texttt{4 and 9} task. We also generated \textbf{saliency maps} to visualize which input pixels\cite{ref7} influenced predictions, both before and after pruning.
\end{enumerate}

All experiments were implemented using Python, with core computations and modeling relying on NumPy \cite{ref8}, Scikit-learn \cite{ref9}, and Matplotlib \cite{ref10} for visualization.

\section{Result}
\subsection{Analysis of Learning Rate and Hidden Size Effects}
To select appropriate training settings for downstream experiments, we conducted a grid search over learning rates and hidden layer sizes. Each model was trained for 100 epochs, and metrics were recorded. 

The tuning was conducted on the digit pair \texttt{4 and 9}, selected as a challenging binary task from the MNIST dataset.

\begin{figure}[htbp]
\centering
\includegraphics[width=0.8\textwidth]{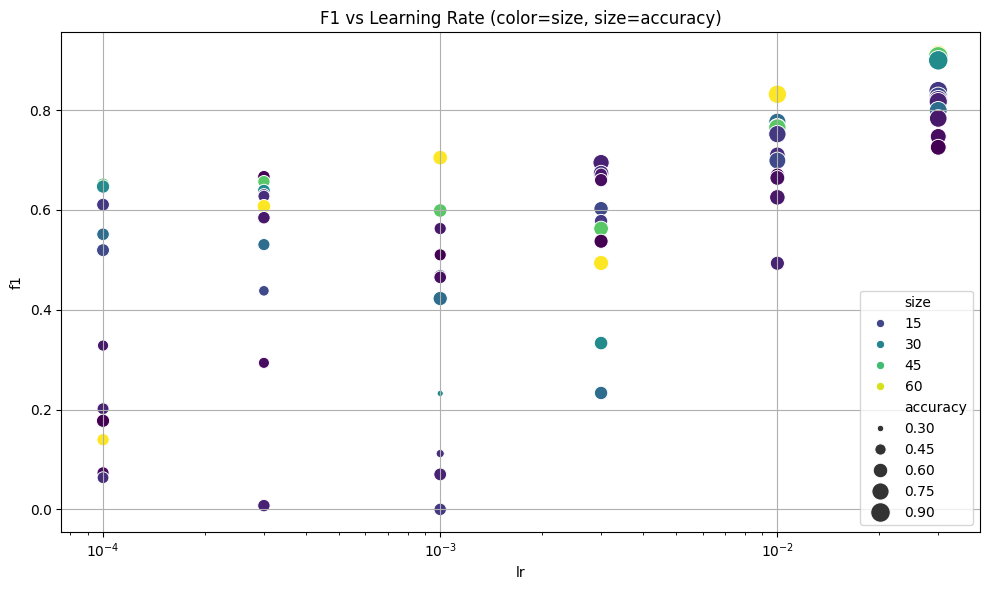} % Replace with actual image path
\caption{F1 vs Learning Rate}
\label{fig:learning_rate__f1_vs_lr}
\end{figure}

The grid search highlighted the critical role of learning rate in optimizing performance. Models trained with very low learning rates (e.g.\ $10^{-4}$) under-performed, likely due to slow convergence and inability to escape poor local minima. In contrast, models with higher learning rates (e.g.\ $10^{-2}$ to $3 \times 10^{-2}$) reached higher optima, as evident by consistent advancements in AUC and F1 scores.

We also observed that score advancements started diminishing past $32$ hidden units. While performance gains continued up to $64$ neurons, the improvements were marginal compared to the model complexity and training time.
For consistency, and because $3 \times 10^{-2}$ also showed strong performance on simpler tasks, we used it for all experiments.
\clearpage

\begin{figure}[htbp]
\centering
\includegraphics[width=0.6\textwidth]{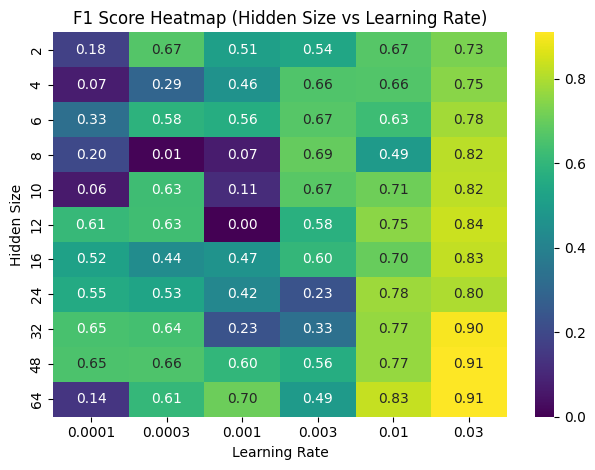} % Replace with actual image path
\caption{Hidden Size vs Learning Rate Heatmap of F1 score over different learning rates and hidden size.}
\label{fig:learning_rate__f1_score_heatmap}
\end{figure}

Moreover, mid-ranging sizes showed instability at certain hidden size combinations, indicating sensitivity to parameter initialization and gradient dynamics in under-powered models. This reinforces the need for adequate model capacity to exploit aggressive learning rates effectively.

However, a learning rate of $3 \times 10^{-2}$ has consistently produced the best results with every hidden size. These findings justify our final selection of a learning rate of $3 \times 10^{-2}$, used in subsequent experiments, yielding the highest F1 scores without signs of overfitting or instability.

\subsection{Determining Minimum Capacity Across Pairs}
To determine the optimal capacity for each pair in the set $\{(0,1),\ (1,7),\ (3,8),\ (4,9),\ (5,6)\}$, we trained separate models using various hidden sizes: $\{2,\ 4,\ 6,\ 8,\ 10,\ 12,\ 16,\ 24,\ 32,\ 48,\ 64\}$. For each model, the best F1 score was selected and plotted with error bars indicating the mean and standard deviation across multiple runs.

\begin{figure}[htbp]
\centering
\includegraphics[width=0.8\textwidth]{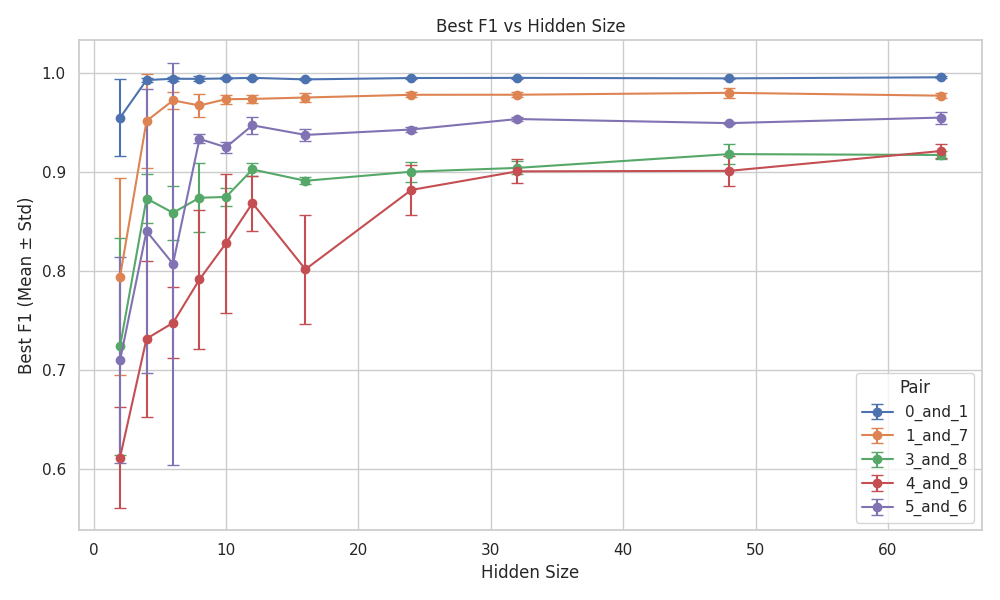}
\caption{Mean F1 Score vs. Hidden Size across five binary classification tasks. The vertical stratification of the curves corresponds to task difficulty, while the error bars represent the standard deviation across three random seeds.}
\label{fig:f1_vs_hidden_size}
\end{figure}

By analyzing F1 scores and loss curves, we established a performance criterion based on achieving both a high F1 score ($> 0.90$) and stable convergence across multiple seeds. The results in Fig. 3. clearly demonstrate that the model complexity required is directly proportional to task complexity.

\subsubsection{Low Difficulty Tasks}
For pair \texttt{0 and 1}, a minimal model with size = 4 (see Orange Curve in Fig. A.1) achieves a near perfect score with a highly stable F1 score and loss $\approx 0.1$. Increasing capacity beyond this resulted in high complexity while yielding marginal decrease in loss.

\subsubsection{Medium Difficulty Tasks}
For pairs \texttt{1 and 7} and \texttt{5 and 6}, the capacity required to distinguish increases significantly as the visual ambiguity became more prominent. Smaller models with size $<$ 8 could sometimes achieve high scores, but were riddled with significant training instability (wide variance bands and performance dips). This wasn't noticeable with models having size $\geq 10$ (see Fig. A.2--A.3).

\subsubsection{High Difficulty Tasks}
Visually similar classes (e.g., \texttt{4 and 9}) were the most demanding. Models with size $<$ 12 clearly under-performed, showing an inability to learn meaningful features, resulting in low F1 scores ($< 0.90$) and stagnated losses. To reliably classify these pairs, a capacity of size 12 to 24 was necessary.

Furthermore, we observed that models operating at their \textbf{critical capacity} limit consistently displayed signatures of instability, including wide performance variance across seeds and erratic learning curves. This instability serves as an empirical marker that the model is under-capacitated. Interestingly, we also noted minor performance inversions (e.g., size = 12 outperforming size = 16 on the \texttt{4 and 9} task), highlighting the non-trivial nature of the optimization landscape even in these small networks.

The detailed learning dynamics for each individual run, showing convergence over 100 epochs, are available for review in the Appendix (Fig. A.1--A.5).

\subsection{Extreme Network Robustness to Magnitude Pruning}
To investigate network redundancy and structure of learned parameters, we conducted one-shot magnitude based pruning. We iteratively pruned weights with the lowest magnitudes and evaluated performance on the validation set without retraining. Fig. 4 reveals the trained models posses highly sparse solutions and are extremely robust to pruning.

\begin{figure}[htbp]
\centering
\includegraphics[width=0.8\textwidth]{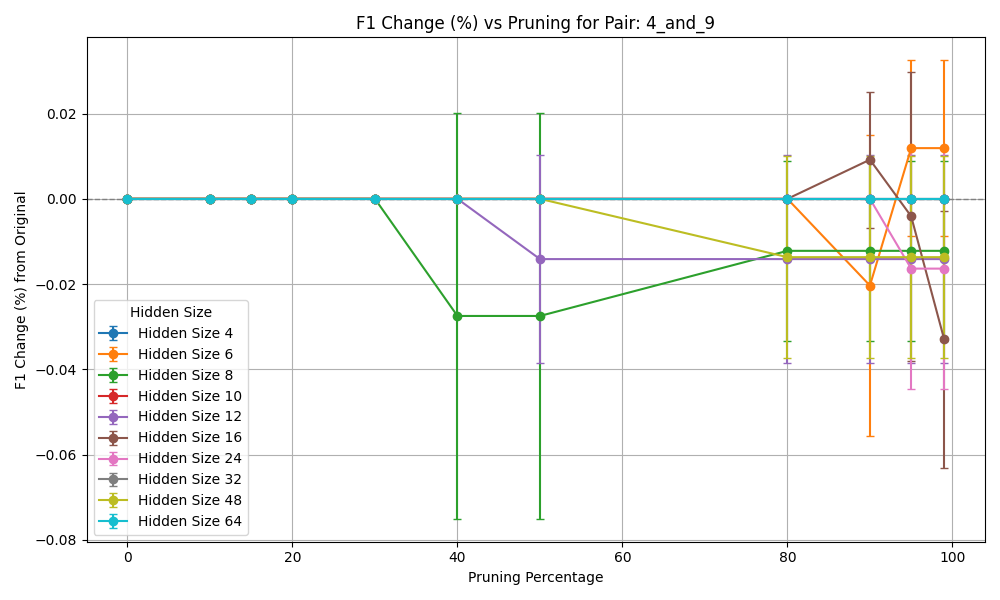}
\caption{Percent Change in F1 score after pruning for Pair 4 and 9, displayed as a representative of dataset}
\label{fig:f1_magnitude_pruning_4_and_9}
\end{figure}

Across all tasks, models demonstrated extreme resilience to high levels of sparsity. For instance, most models showed no discernible drop ($\approx 10^{-5}$) even with 99\% pruning for the simple \texttt{0 and 1} task (Fig. B.1). This strongly indicates that the network learns a highly redundant representation, where only a tiny fraction of parameters are critical for solving a simple problem. Even for a complex task like \texttt{4 and 9}, the drop is marginal in levels of $10^{-2}$.

Interestingly, we also found evidence of \textbf{pruning acting as a regularizer}. In several cases, particularly for harder tasks, performance slightly \textit{increased} at moderate to high sparsity levels. This suggests that pruning can improve generalization by eliminating spurious, low-magnitude weights that may have contributed to minor overfitting on the training set.

Significant performance degradation was rare and graceful, primarily occurring at extreme sparsity ($>95\%$) and affecting models that were already operating at their critical capacity limit. This provides converging evidence that the size of the essential, high-magnitude subnetwork is directly related to task complexity. Simpler tasks require a smaller, well-performing subset, making them more resilient to pruning.

\subsection{Analysis of Dead Neurons}
To understand the structural impact of pruning on the hidden layer, we analyzed the prevalence of \textbf{dead neurons}, neurons that output zero for all inputs in the validation set. We investigated this both before and after applying 95\% magnitude pruning.

\begin{table}[h]
\centering
\caption{Number of dead neurons before and after 95\% magnitude pruning for selected models.}
\begin{tabular}{|l|l|l|l|}
\specialrule{1.5pt}{0pt}{0pt}
\textbf{Task Pair} & \textbf{Model Size} & \textbf{Dead Neurons (Original)} & \textbf{Dead Neurons (After Pruning)} \\
\midrule
0 and 1 & 12 & 0 & 0 \\
3 and 8 & 16 & 0 & 0 \\
4 and 9 & 24 & 0 & 0 \\
\specialrule{1.5pt}{0pt}{0pt}
\end{tabular}
\label{tab:dead_neurons}
\end{table}

Strikingly, our findings showed that all tested models had \textbf{zero dead neurons} in their original, unpruned state. This indicates an efficient and \textbf{healthy} training process where all allocated neurons participate in the network's function.

Even more remarkably, the number of dead neurons remained zero after 95\% of the weights were removed. This key result implies that the surviving high-performing subnetwork is not localized to a few critical neurons. Instead, the essential connections are \textbf{distributed across the entire hidden layer}, ensuring every neuron remains an active component of the computational graph. This demonstrates that the network's learned sparsity exists at the \textbf{weight level, not the neuron level}.

\subsection{Visualizing Clusters}
To visually inspect the quality of the learned representations, we used t-SNE to project the hidden layer activations of the size = 24 model onto a 2D space (Fig. 5). The analysis was performed on the challenging \texttt{4 and 9} task, both before and after 95\% magnitude pruning. We chose this size as it was discovered to be adequate for the classification.

\begin{figure}[htbp]
\centering
\includegraphics[width=0.8\textwidth]{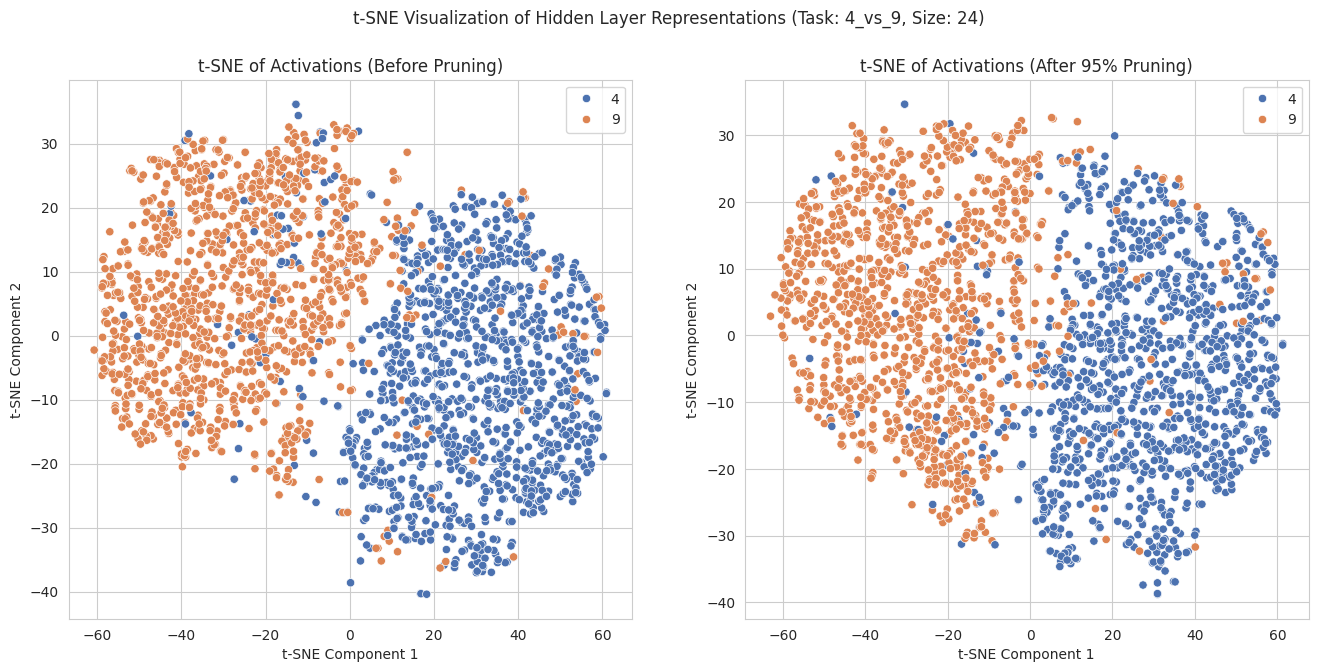}
\caption{t-SNE visualization of hidden layer before and after 95\% pruning for pair 4 and 9 and size = 24}
\label{fig:t-sne}
\end{figure}

The visualization of the original network (left) shows two distinct, well-separated clusters, corresponding to the two digit classes. This confirms that the model learned a powerful internal representation for classification.

Critically, the t-SNE visualization of the network after 95\% of its weights were pruned (right) is qualitatively identical. The class clusters remain just as compact and linearly separable. This provides strong visual evidence that the \textbf{surviving sparse subnetwork} preserves the rich geometric structure of the original learned feature space, explaining why its performance remains fully intact.

\subsection{Robustness to Input Correction}
We investigated the effect of model capacity on robustness against two types of input corruption: additive Gaussian noise and random pixel occlusion. We evaluated a model of \textbf{suitable} capacity (size = 24) and an overparameterized model (size = 64) on the challenging \texttt{4 and 9} task.

\begin{figure}[htbp]
\centering
\includegraphics[width=0.6\textwidth]{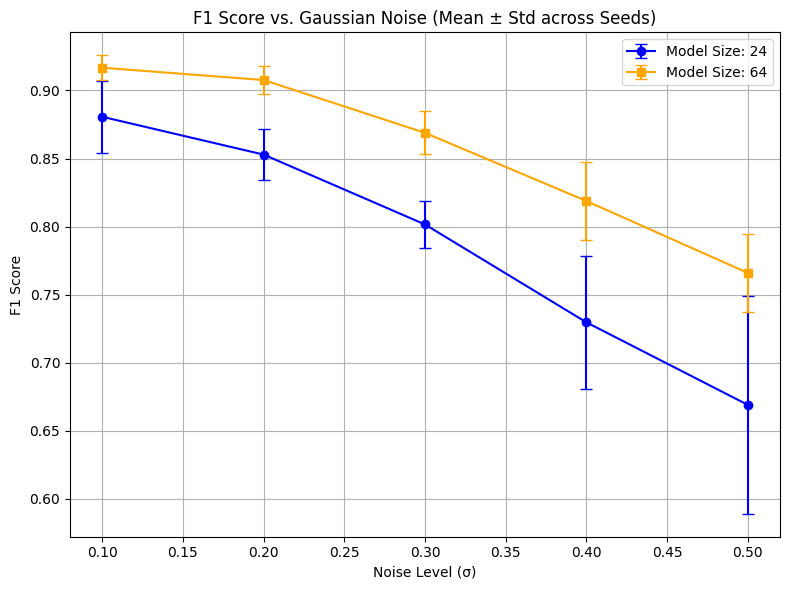}
\caption{F1 Score for model of size 24 and 64 post applying gaussian noise}
\label{fig:gaussian}
\end{figure}

As shown in \textbf{Fig. 6}, the over-parameterized size = 64 model consistently maintained a higher F1 score than the size = 24 model across all tested levels of Gaussian noise.

\begin{figure}[htbp]
\centering
\includegraphics[width=0.6\textwidth]{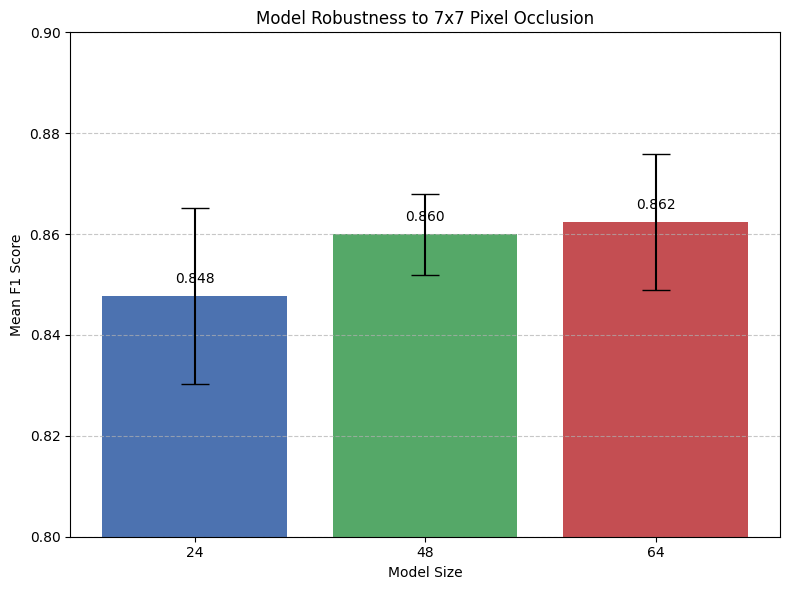}
\caption{Model Robustness to 7x7 Pixel Occlusion}
\label{fig:occlusion}
\end{figure}

We observed the same trend in the occlusion experiment, where a random $7 \times 7$ patch of the input image was set to zero. On the occluded dataset, performance consistently improved with capacity, with the size = 64 model achieving the highest mean F1 score (0.862), followed by the size = 48 model (0.860), and the size = 24 model (0.847). Interestingly, we observed diminishing returns, where the robustness gain from size = 48 to size = 64 was marginal.

These converging results suggest that the excess capacity was used to learn a more \textbf{redundant and resilient representation}, making larger models more robust to various forms of input corruption.

\subsection{Interpretability via Saliency Mapping}
To gain a deeper, pixel-level insight into the model's decision-making process, we generated saliency maps, a foundational method for feature attribution, in a field that has since developed more advanced axiomatic approaches\cite{ref16}. This analysis allows us to move beyond performance metrics and understand the model's internal ``attentional'' strategy. We focused on the challenging \texttt{4 and 9} task with the size = 24 model, examining its behavior under various conditions (Fig. 8).

\begin{figure}[htbp]
\centering
\begin{subfigure}[b]{0.45\textwidth}
    \centering
    \includegraphics[width=\textwidth]{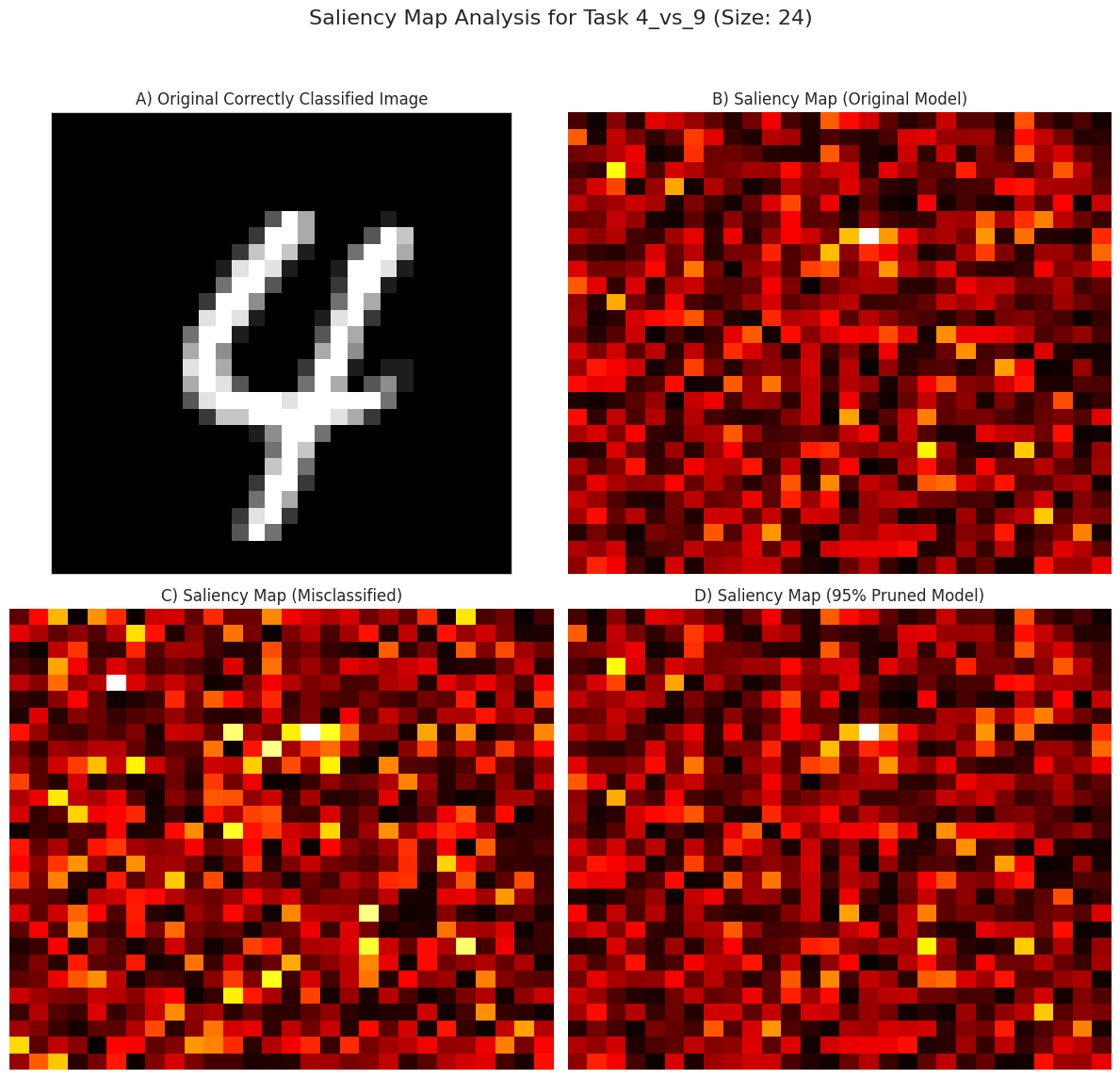}
    \caption{Classification of image representing digit 4}
    \label{fig:4}
\end{subfigure}
\hspace{0.5cm}
\begin{subfigure}[b]{0.45\textwidth}
    \centering
    \includegraphics[width=\textwidth]{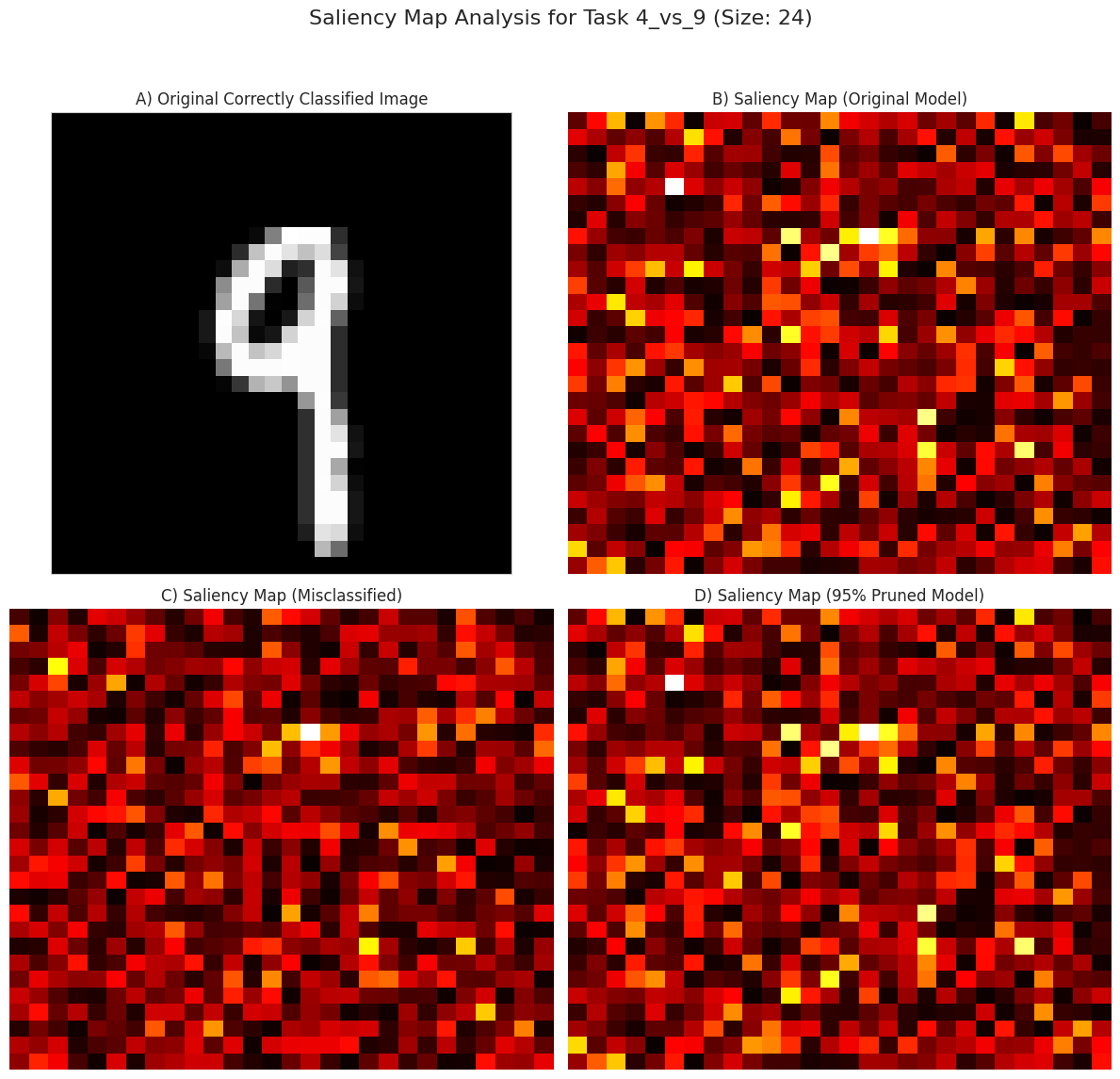}
    \caption{Classification of image representing digit 4}
    \label{fig:9}
\end{subfigure}
\caption{Saliency map for digit '4' and '9'}
\label{fig:saliency}
\end{figure}

\subsubsection{Baseline: Correct Classification Reveals Learned Features}
First, we analyzed the model's reasoning for a correctly classified image. For a \texttt{4} (Fig. 8.a), the saliency map shows high importance (bright pixels) concentrated on the digit's defining features: the primary vertical stem and, critically, the horizontal cross-bar. Similarly, for a correctly classified \texttt{9}, the map highlights the top loop and long stem. This confirms that the model has learned a semantically meaningful feature detector, grounding its decisions in the correct, human-interpretable parts of the image.

\subsubsection{Failure Analysis: A "Smoking Gun" for Misclassification}
The mechanism of model failure is clearly revealed when analyzing a misclassified example. For a \texttt{4} that was incorrectly classified as a \texttt{9} (Fig. 8.a), the saliency map shows a critical shift in attention. The model assigned low importance to the horizontal cross-bar, effectively ignoring the single most important feature that distinguishes a \texttt{4} from a \texttt{9}. Instead, it focused on the upper part of the vertical stem and the adjacent empty space, features that are ambiguous and also present in a \texttt{9}. This provides direct visual evidence that misclassification is directly tied to a failure in correct feature attribution.

\subsubsection{The Pruned Sub-network Preserves Reasoning}
Finally, we investigated whether this reasoning process was preserved in the sparse subnetwork. The saliency map for the 95\% pruned model is remarkably consistent with the original model's map for a correct classification. It focuses on the same essential strokes, demonstrating that the pruned subnetwork doesn't just replicate the accuracy, but also the \textbf{learned attentional strategy} of its parent network.

\section{Discussion}
Our work provides a systematic investigation into the foundational properties of low-capacity neural networks, using a controlled framework to link task complexity to model behavior. The central finding is a direct, quantitative relationship between the visual difficulty of a classification task and the minimum model capacity required for effective generalization. By first establishing these capacity baselines, we were able to conduct a deeper analysis into the sparsity and robustness of the resulting networks, leading to several key insights.

Our analysis of network pruning revealed more than just high performance at extreme sparsity. The t-SNE visualizations and dead neuron analysis demonstrate that these sparse subnetworks are structurally sound, preserving the geometric separability of the learned representations without collapsing. Furthermore, the saliency maps provide compelling evidence that these subnetworks retain the core "reasoning" of their parent models, focusing on the same semantically relevant features. This suggests that the sparse solution found through pruning is not an alien logic but a true, efficient microcosm of the original network.

These findings present an interesting paradox regarding the role of overparameterization. On one hand, we show that trained models contain highly sparse solutions, suggesting most parameters are redundant post-training. On the other hand, our robustness experiments demonstrate that models that started with more excess capacity were significantly more resilient to input noise and occlusion. This implies that overparameterization, while perhaps not essential for the final sparse architecture, may be a crucial ingredient during the training process itself, a phenomenon explored in work on the double descent of generalization error\cite{ref12}. It may provide the necessary flexibility for the optimizer to discover a more general and robust set of features that can withstand real-world corruption.

Finally, we acknowledge the limitations of this study. Our experiments were conducted exclusively on the MNIST dataset with a simple fully-connected architecture. While this controlled setting provides clear insights, future work is needed to validate whether these specific relationships hold for more complex architectures like Convolutional Neural Networks and on more challenging, real-world datasets.

\section{Conclusion and Future Work}
In this work, we developed a structured framework to study the fundamental behavior of low-capacity neural networks. This was done through a suite of binary classification tasks designed with gradually increasing difficulty. Our empirical findings can be summarized as follows:

\begin{itemize}
    \item \textbf{Capacity scales with task complexity:} The minimum number of hidden units required for generalization grows as the visual similarity between digit classes increases.

    \item \textbf{Sparse subnetworks retain core function:} Even after pruning up to 99\% of weights, networks often contain subnetworks that preserve both performance and their internal reasoning patterns.

    \item \textbf{Overparameterized networks often rely on redundancy:} They seem essential for learning internal representations that remain stable under input noise or corruption.
\end{itemize}

Taken together, these results highlight how capacity, sparsity, and robustness are tightly linked, even in the most basic network settings.

Looking forward, several future directions appear promising:

\begin{itemize}
    \item \textbf{Neuron-level interpretability:} Investigating the internal structure of these models by:
    \begin{itemize}
        \item Computing \textbf{mutual information} between individual neuron activations and class labels.
        \item Defining a \textbf{neuron specialization score} to quantify task-specific adaptation within hidden units.
    \end{itemize}

    \item \textbf{Saliency map comparisons:} Analyzing how model attention shifts under different conditions by:
    \begin{itemize}
        \item Comparing saliency patterns across architectures.
        \item Evaluating changes under varying levels of input corruption.
    \end{itemize}
\end{itemize}

These directions may help reveal \textbf{how information is structured and manipulated} within low-capacity networks, an area still underexplored but critical for deploying reliable models in constrained or sensitive environments.

\clearpage
\bibliographystyle{IEEEtran}  % or use 'plain', 'unsrt', etc.
\bibliography{references}

\clearpage
\appendix
\section*{Appendix}
\addcontentsline{toc}{section}{Appendix} % only if you have a TOC

\subsection*{A. Detailed Training Curves}
\renewcommand{\thefigure}{A.\arabic{figure}}
\setcounter{figure}{0}

This section provides the detailed F1 score and loss curves for each of the five binary classification tasks across the full range of hidden layer sizes. The plots show the learning dynamics over 100 epochs, with performance averaged across three random seeds.

\begin{figure}[htbp]
\centering
\begin{subfigure}[b]{0.45\textwidth}
    \centering
    \includegraphics[width=\textwidth]{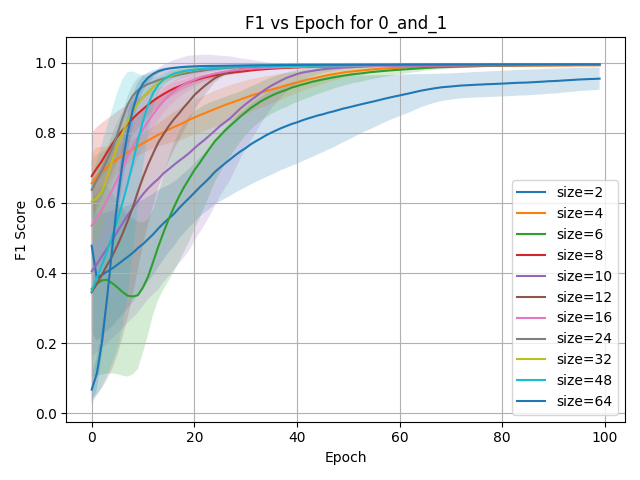}
    \caption{}
    \label{fig:f1_vs_epoch_0_and_1}
\end{subfigure}
\hspace{0.5cm}
\begin{subfigure}[b]{0.45\textwidth}
    \centering
    \includegraphics[width=\textwidth]{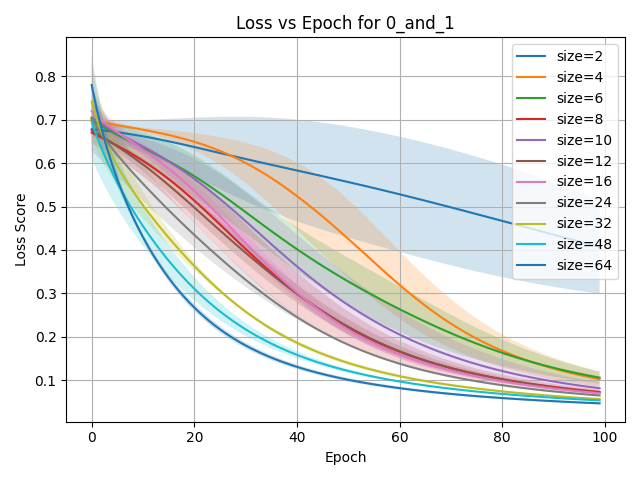}
    \caption{}
    \label{fig:loss_vs_epoch_0_and_1}
\end{subfigure}
\caption{Training dynamics for the 0 vs. 1 task. (Left) F1 score vs. epoch. (Right) Loss vs. epoch. Each colored line represents a different hidden layer size. The shaded regions indicate the standard deviation across three random seeds, visualizing training stability.}
\label{fig:loss_and_f1_vs_epoch_0_and_1}
\end{figure}

\begin{figure}[htbp]
\centering
\begin{subfigure}[b]{0.45\textwidth}
    \centering
    \includegraphics[width=\textwidth]{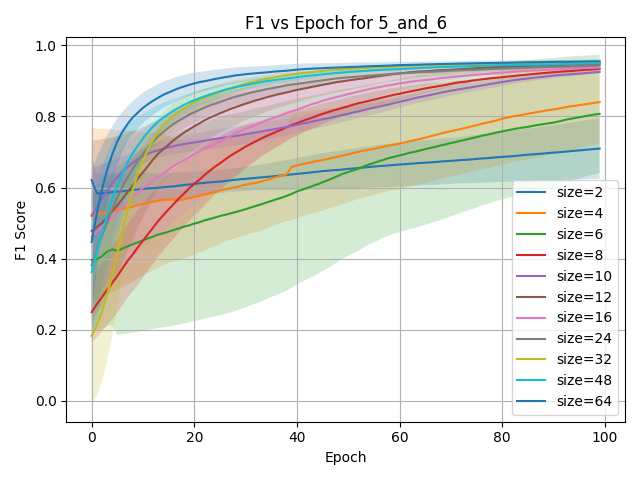}
    \caption{}
    \label{fig:f1_vs_epoch_5_and_6}
\end{subfigure}
\hspace{0.5cm}
\begin{subfigure}[b]{0.45\textwidth}
    \centering
    \includegraphics[width=\textwidth]{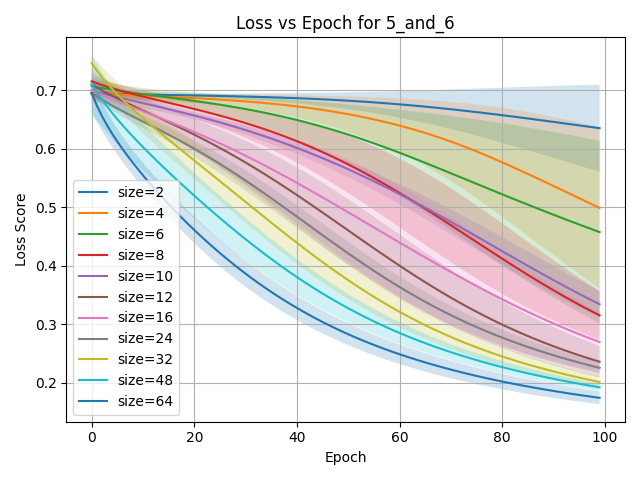}
    \caption{}
    \label{fig:loss_vs_epoch_5_and_6}
\end{subfigure}
\caption{Training dynamics for the 5 and 6 task. (Left) F1 score vs. epoch. (Right) Loss vs. epoch.}
\label{fig:f1_and_loss_vs_epoch_5_and_6}
\end{figure}

\begin{figure}[htbp]
\centering
\begin{subfigure}[b]{0.45\textwidth}
    \centering
    \includegraphics[width=\textwidth]{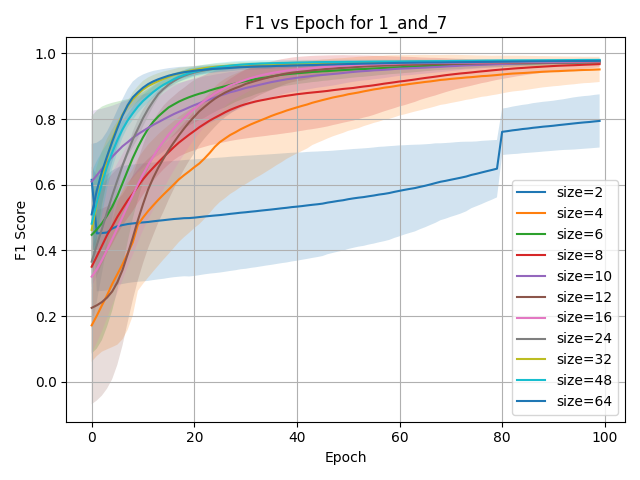}
    \caption{}
    \label{fig:f1_vs_epoch_1_and_7}
\end{subfigure}
\hspace{0.5cm}
\begin{subfigure}[b]{0.45\textwidth}
    \centering
    \includegraphics[width=\textwidth]{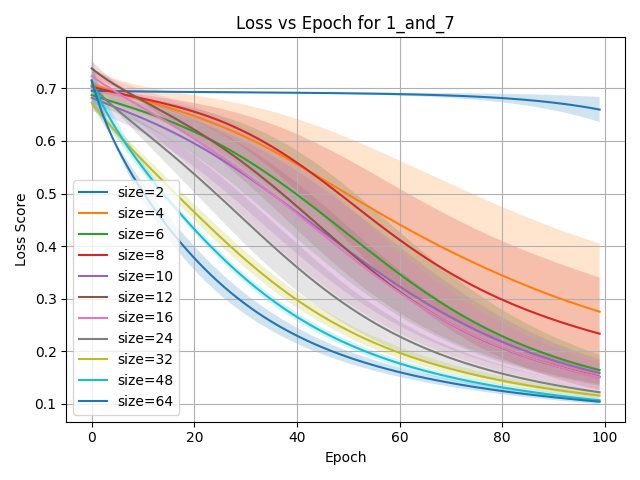}
    \caption{}
    \label{fig:loss_vs_epoch_1_and_7}
\end{subfigure}
\caption{Training dynamics for the 1 and 7 task. (Left) F1 score vs. epoch. (Right) Loss vs. epoch.}
\label{fig:f1_and_loss_vs_epoch_1_and_7}
\end{figure}

\begin{figure}[htbp]
\centering
\begin{subfigure}[b]{0.45\textwidth}
    \centering
    \includegraphics[width=\textwidth]{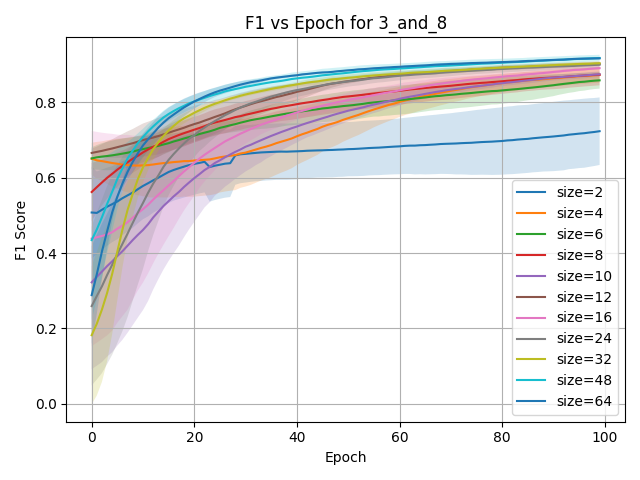}
    \caption{}
    \label{fig:f1_vs_epoch_3_and_8}
\end{subfigure}
\hspace{0.5cm}
\begin{subfigure}[b]{0.45\textwidth}
    \centering
    \includegraphics[width=\textwidth]{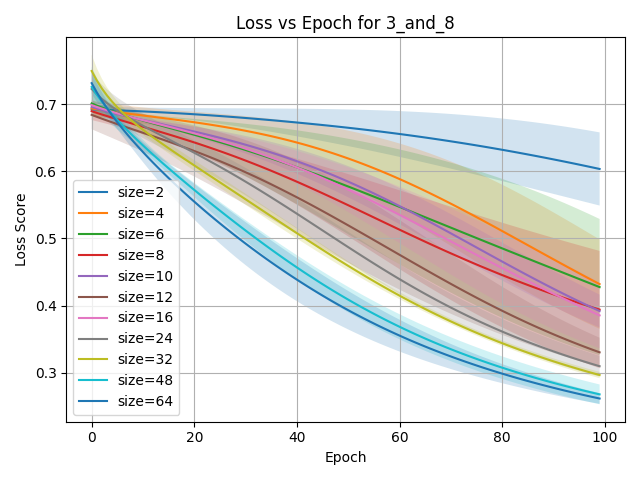}
    \caption{}
    \label{fig:loss_vs_epoch_3_and_8}
\end{subfigure}
\caption{Training dynamics for the 3 and 8 task. (Left) F1 score vs. epoch. (Right) Loss vs. epoch.}
\label{fig:f1_and_loss_vs_epoch_3_and_8}
\end{figure}

\begin{figure}[htbp]
\centering
\begin{subfigure}[b]{0.45\textwidth}
    \centering
    \includegraphics[width=\textwidth]{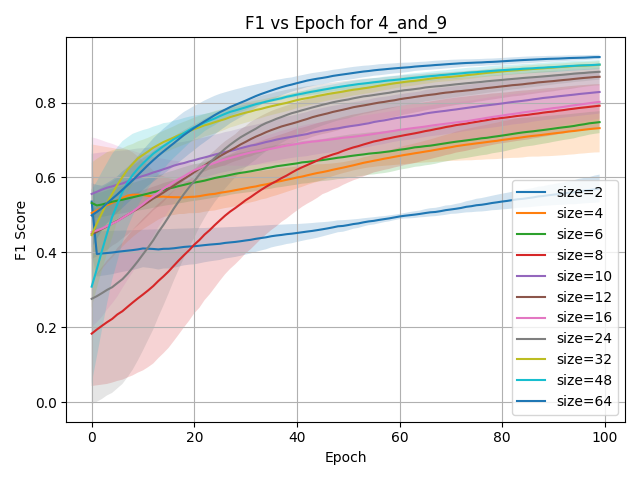}
    \caption{}
    \label{fig:f1_vs_epoch_4_and_9}
\end{subfigure}
\hspace{0.5cm}
\begin{subfigure}[b]{0.45\textwidth}
    \centering
    \includegraphics[width=\textwidth]{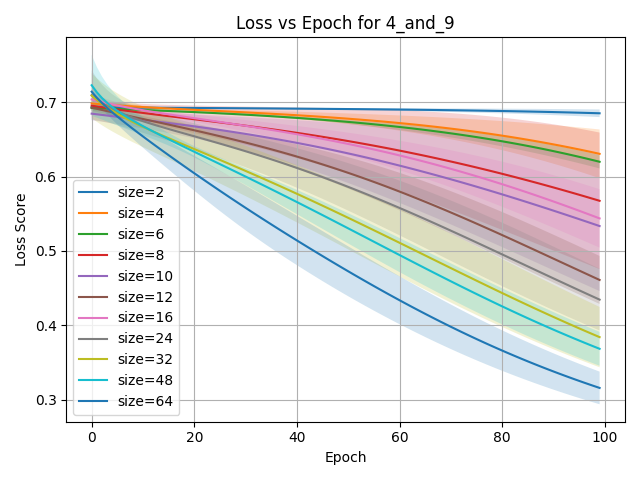}
    \caption{}
    \label{fig:loss_vs_epoch_4_and_9}
\end{subfigure}
\caption{Training dynamics for the 4 and 9 task. (Left) F1 score vs. epoch. (Right) Loss vs. epoch.}
\label{fig:f1_and_loss_vs_epoch_4_and_9}
\end{figure}

\clearpage
\subsection*{B. Pruning Curves}
\renewcommand{\thefigure}{B.\arabic{figure}}
\setcounter{figure}{0}

\begin{figure}[htbp]
\centering
\includegraphics[width=0.8\textwidth]{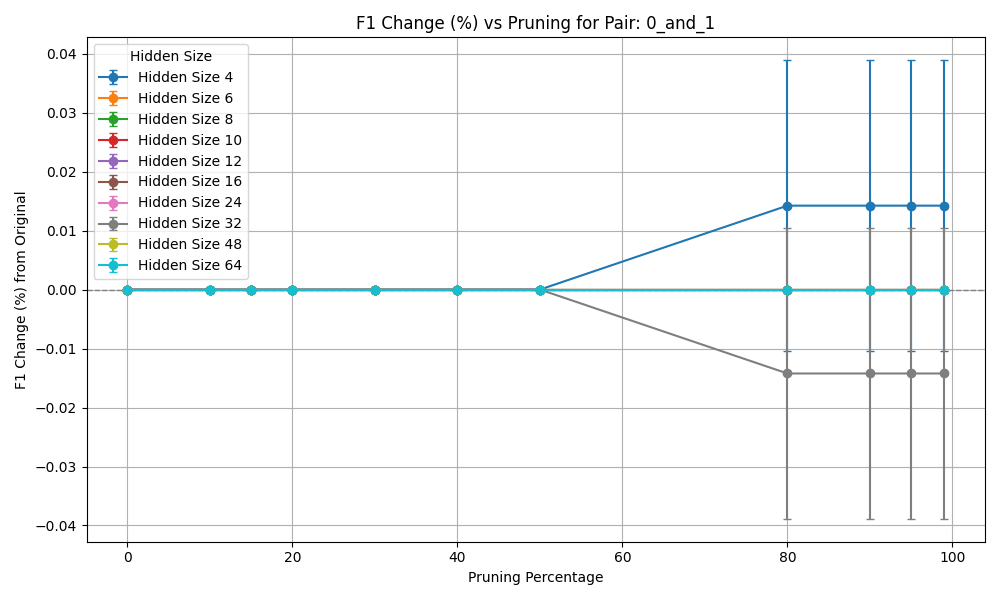}
\caption{Percentage change in F1 Score after pruning for pair 0 and 1}
\label{fig:f1_magnitude_pruning_0_and_1}
\end{figure}

\begin{figure}[htbp]
\centering
\includegraphics[width=0.8\textwidth]{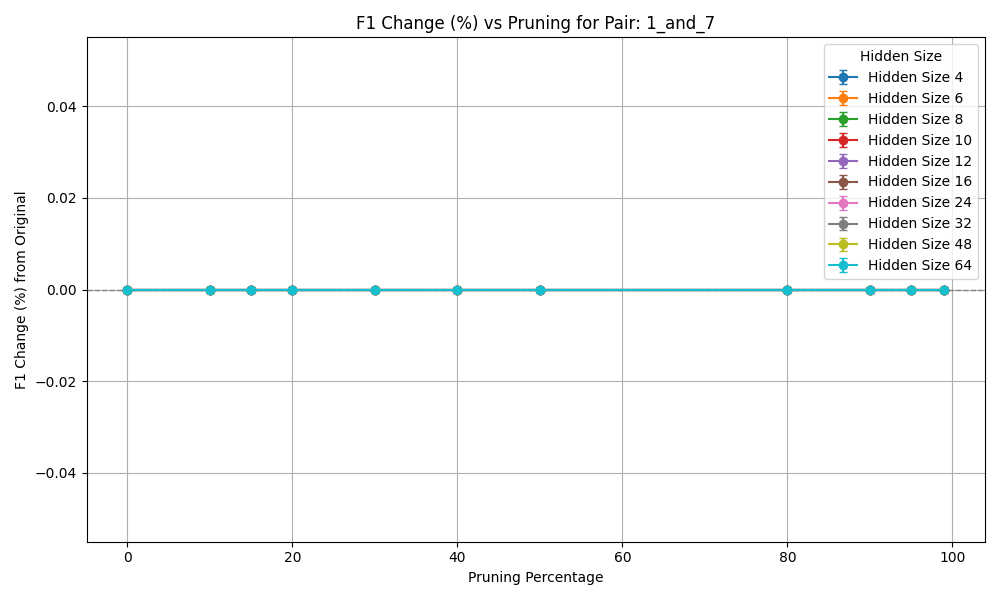}
\caption{Percentage change in F1 Score after pruning for pair 1 and 7}
\label{fig:f1_magnitude_pruning_1_and_7}
\end{figure}

\begin{figure}[htbp]
\centering
\includegraphics[width=0.8\textwidth]{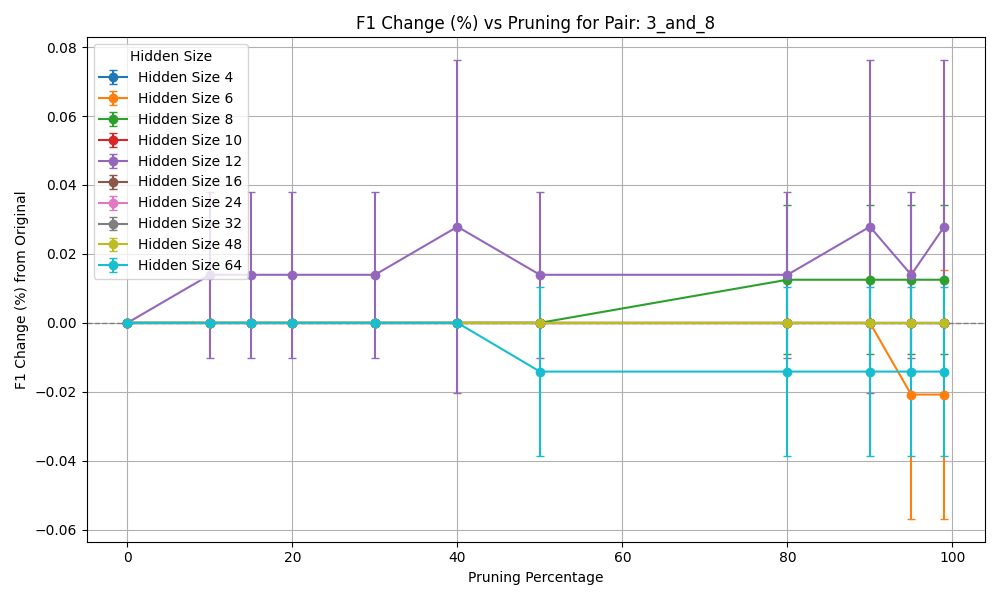}
\caption{Percentage change in F1 Score after pruning for pair 3 and 8}
\label{fig:f1_magnitude_pruning_3_and_8}
\end{figure}

\begin{figure}[htbp]
\centering
\includegraphics[width=0.8\textwidth]{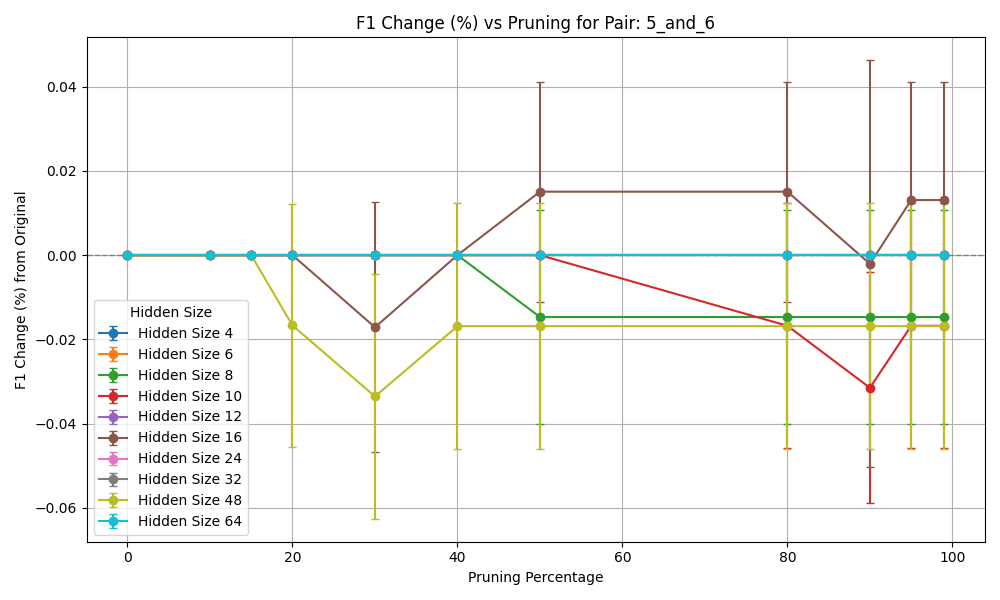}
\caption{Percentage change in F1 Score after pruning for pair 5 and 6}
\label{fig:f1_magnitude_pruning_5_and_6}
\end{figure}

\clearpage
\subsection*{C. Core Function Implementations}
\renewcommand{\thefigure}{C.\arabic{figure}}
\setcounter{figure}{0}
For reproducibility, we provide the core Python functions used for our pruning and interpretability analyses.

\textbf{C.1. Python implementation for one-shot magnitude pruning}
\begin{lstlisting}[language=Python]
def _apply_magnitude_pruning(self):
    # Prune w_1
    flat_w1 = self.w_1.flatten()
    k1 = int(self.prune_prob * flat_w1.size)
    if k1 > 0:
        threshold1 = np.partition(np.abs(flat_w1), k1 - 1)[k1 - 1]
        self.w_1[np.abs(self.w_1) <= threshold1] = 0
    # Prune w_2
    flat_w2 = self.w_2.flatten()
    k2 = int(self.prune_prob * flat_w2.size)
    if k2 > 0:
        threshold2 = np.partition(np.abs(flat_w2), k2 - 1)[k2 - 1]
        self.w_2[np.abs(self.w_2) <= threshold2] = 0
\end{lstlisting}

\textbf{C.2. Python implementation for Saliency Maps}
\begin{lstlisting}[language=Python]
def create_saliency_map(model, image_vector, epsilon=0.01):
    baseline_score = model.predict(image_vector.reshape(1, -1))[0]
    saliency_values = np.zeros_like(image_vector)
    for i in range(image_vector.shape[0]):
        perturbed_image = np.copy(image_vector)
        perturbed_image[i] += epsilon
        new_score = model.predict(perturbed_image.reshape(1, -1))[0]
        saliency_values[i] = np.abs(new_score - baseline_score)
    return saliency_values.reshape(28, 28)
\end{lstlisting}

\subsection*{D. Examples of Input Corruption}
\renewcommand{\thefigure}{D.\arabic{figure}}
\setcounter{figure}{0}

\begin{figure}[htbp]
\centering
\includegraphics[width=0.8\textwidth]{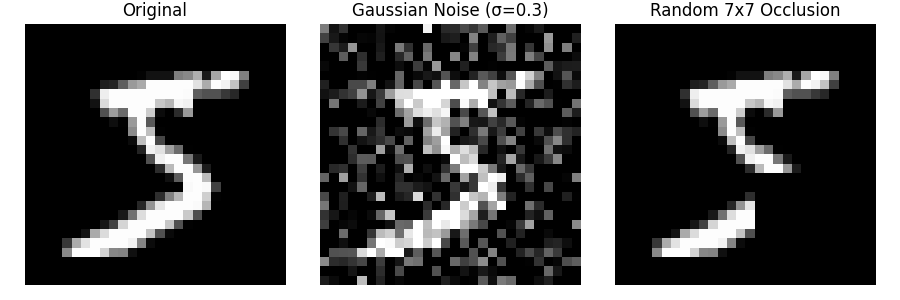}
\caption{(a) Original image from the MNIST dataset. (b) The same image with additive Gaussian noise ($\sigma = 0.3$). (c) The same image with a random 7x7 occlusion patch.}
\label{fig:noise_and_occlusion}
\end{figure}
\end{document}